# Attention-GAN for Object Transfiguration in Wild Images

Xinyuan Chen[1,2], Chang Xu[3], Xiaokang Yang[1], and Dacheng Tao[3]

[1]Shanghai Jiao Tong University
[2]University of Technology, Sydney
[3]University of Sydney

**Abstract**

This paper studies the object transfiguration problem in wild images. The generative network in classical GANs for object transfiguration often undertakes a dual responsibility: to detect the objects of interests and to convert the object from source domain to target domain. In contrast, we decompose the generative network into two separat networks, each of which is only dedicated to one particular sub-task. The attention network predicts spatial attention maps of images, and the transformation network focuses on translating objects. Attention maps produced by attention network are encouraged to be sparse, so that major attention can be paid to objects of interests. No matter before or after object transfiguration, attention maps should remain constant. In addition, learning attention network can receive more instructions, given the available segmentation annotations of images. Experimental results demonstrate the necessity of investigating attention in object transfiguration, and that the proposed algorithm can learn accurate attention to improve quality of generated images.

## 1 Introduction

The task of image-to-image translation aims to translate images from a source domain to another target domain, e.g., greyscale to color and image to semantic label. A lot of researches on image-to-image translation have been produced in the supervised setting, where ground truths in the target domain are available. [1] learns a parametric translation function using CNNs by minimizing the discrepancy between generated images and the corresponding target images. [2] uses conditional GANs to learn a mapping from input to output images. Similar ideas have been applied to various tasks such as generating photographs from sketch or from semantic layout [3, 4], and image super-resolution [5].

To achieve image-to-image translation in the absence of paired examples, a series of works has emerged by combining classical adversarial training [6] with different carefully designed constraints, e.g., circularity constraint [7, 8, 9], $f$-consistency constraint [10], and distance constraints [11]. Although there is no paired data, these constraints are applied to establish the connections between two domains so that meaningful analogs are obtained. Circularity constraint [7, 8, 9] requires a sample from one domain to the other that can be mapped back to produce the original sample. $f$-consistency requires both input and output in each domain that should be consistent with each other in intermediate space of a neural network. [11] learns the image translation mapping in a one-sided unsupervised way by enforcing high cross-domain correlation between matching pairwise distances computed in source and target domains.

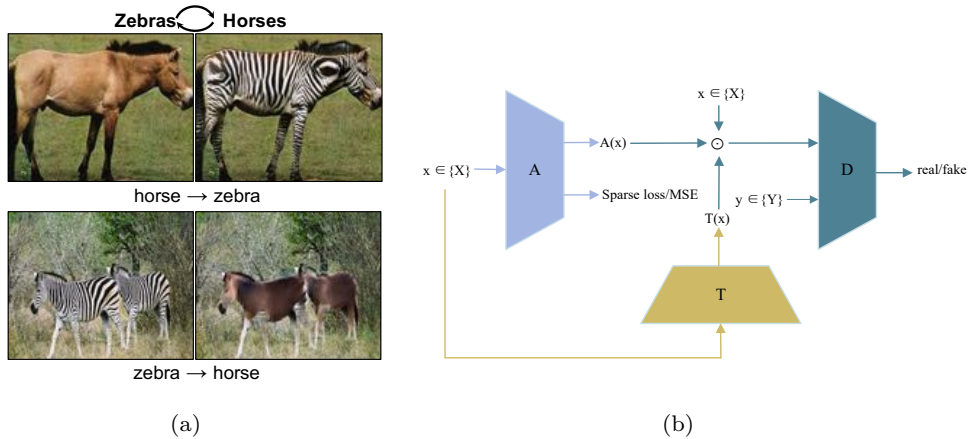

Figure 1: (a): Object transfiguration of horse ↔ zebra. (b): An illustration of Attention-GAN. $A, T, D$ respectively represent the attention network, the transformation network and the discriminative network. Sparse loss denotes the sparse regularization for the predicted attention map. MSE denotes mean square error loss for supervised learning. $A(x)$ denotes the attention map predicted by the attention network. $T(x)$ denotes the transformed images. $\odot$ denotes the layered operation.

Object transfiguration is a special task in the image-to-image translation problem. Instead of taking the image as a whole to accomplish the transformation, object transfiguration aims to transform a particular type of object in an image to another type of object without influencing the background regions. For example, in the top line of Figure 1(a), horses in the image are transformed into zebras, and zebras are transformed into horses, but the grassland and the trees are expected to be constant. Existing methods [7, 11] are used to tackle object transfiguration as a general image-to-image task, without investigating unique insights of the problem. In such a one-shot generation, a generative network actually takes two distinct roles: detecting the region of interests and converting object from source domain to target domain. However, incorporating these two functionalities in a single network would confuse the aims of the generative network. During iterations, it could be unclear whether the generative network should improve its detection of the objects of interests or boost its transfiguration of the objects. The quality of generated images is often seriously influenced as a result, e.g. some background regions might be taken into transformation by mistake.

In this paper, we propose an attention-GAN algorithm for the object transfiguration problem. The generative network in classical GANs has been factorized as two separate networks: an attention network to predict where the attention should be paid, and a transformation network that actually carries out the transformation of objects. A sparse constraint is applied over the attention map, so that limited attention energy can be focused on regions of priority rather than spreaded on the whole image at random. A layered operation is adopted to finalize the generated images by combining the transformed objects and the original background regions with the help of the learned sparse attention mask. A discriminative network is employed to distinguish real images from these synthesized images, while attention network and transformation network cooperate to generate synthesized images that can fool the discriminative network. Cycle-consistent loss [7, 8, 9] was adopted to handle unpaired data. Moreover, if segmentation results of images are available, the attention network can be learned in a supervised manner and the performance of the proposed algorithm can be improved accordingly. Experimental results on three object transfiguration tasks, i.e. horse $\leftrightarrow$ zebra, tiger $\leftrightarrow$ leopard, and apple $\leftrightarrow$ orange [12], suggest the advantages of investigating attention in object transfiguration, and the quantitative and the qualitative performance improvement of the proposed algorithm over state-of-the-art methods.

## 2 Related Work

### 2.1 Generative Adversarial Networks

Generative adversarial networks (GANs) [6] have achieved impressive results in image generation [13, 14] by way of a two-player minimax game: a discriminator aims to distinguish the generated images from real images while a generator aims to generate realistic images to fool the discriminator. A series of multi-stage generative models has been proposed to generate more realistic images [15, 16, 17]. [16] proposes composite generative adversarial network (CGAN) that disentangles complicated factors of images by employing multiple generators to generate different parts of the image. The layered recursive GANs [17] learns to generate image background and foregrounds separately and recursively.

GANs have shown a great success on a variety of conditional image generation applications, e.g., image-to-image translation [7, 8, 9], text-to-image generation [18, 19]. Different from the original GANs that generate images from noise variables, conditional GANs synthesize images based on the input information (e.g., category, image and text). In image-to-image translation problem such as sketch to photo, map to aerial photo, day to night etc. [2] investigates conditional adversarial networks for a general solution. After this, [7, 8, 9] introduce cycle consistency loss to solve unpaired image to image translation problem and also conduct experiments on object transfiguration (e.g. horse to zebra and apple to orange). [20] proposes a mask-conditional contrast-GAN architecture to disentangle image background with object semantic changes by exploiting the semantic annotations in both train and test phases. However, it is hard to collect segmentation mask for a large number of images, especially in test phase.

### 2.2 Attention Model in Networks

Motivated by human attention mechanism theories [21], attention mechanism has been successfully introduced in computer vision and natural language processing tasks, e.g. image classification [22, 23, 24], image captioning [25], visual question answering [26], image segmentation [27]. Rather than compressing an entire image or a sequence into a static representation, attention allows the model to focus on the most relevant part of images or features as needed. Mnih et al. [22] propose a recurrent network model that is capable of extracting information from an image or video by adaptively selecting a sequence of regions or locations and only processing the selected regions at high resolution. Bahdanau et al. [28] propose an attention model that softly weights the importance of input words in a source sentence when predicting a target word for machine translation. Following this, Xu et al. [25] and Yao et al. [29] use attention models for image captioning and video captioning respectively. The model automatically learns to fix its gaze on salient objects while generates the corresponding words in the output sequence. In visual question answering, [26] uses the question to choose relevant regions of the images for computing the answer. In image generation, Gregor et al. [30] proposes a generative network combined attention mechanism with a sequential variational auto-encoding framework. The generator attends a smaller region of an input image guided by the ground truth image, and generates a few pixels for an image at a time.

## 3 Preliminaries

In the task of image-to-image translation, we have two domains $X$ and $Y$ with training samples $\{x_i\}_i^N \in X$ and $\{y_i\}_i^N \in Y$. The goal is to learn mapping from one domain to the other $\mathcal{G} : X \to Y$, (e.g. horse→zebra). The discriminator $D_Y$ aims to distinguish real image $y$ from translated images $\mathcal{G}(x)$. On the contrary, the mapping function $\mathcal{G}$ tries to generate images $\mathcal{G}(x)$ that looks similar to images in $Y$ domain to fool the discriminator. The objective of *adversarial loss* in LSGAN [31] is expressed as:

$$\mathcal{L}_{GAN}(\mathcal{G}, D_Y, X, Y) = \mathbb{E}_{y \in Y}\left[(D_Y(y) - 1)^2\right] + \mathbb{E}_{x \in X}\left[D_Y(\mathcal{G}(x))^2\right], \quad (1)$$

The mapping function $\mathcal{F} : Y \to X$, in the same way, tries to fool the discriminator $D_X$:

$$\mathcal{L}_{GAN}(\mathcal{F}, D_X, X, Y) = \mathbb{E}_{x \in X}\left[(D_X(x) - 1)^2\right] + \mathbb{E}_{y \in Y}\left[D_X(\mathcal{F}(y))^2\right]. \quad (2)$$

The discriminators $D_X$ and $D_Y$ try to maximize the loss while mapping functions $\mathcal{G}$ and $\mathcal{F}$ try to minimize the loss. However, a network of sufficient capacity can map the set of input images to any random permutation of images in the target domain. To guarantee that the learned function maps an individual input $x$ to a desired output $y$, the *cycle consistency loss* is proposed to measure the discrepancy occurred when the translated image is brought back to the original image space:

$$\mathcal{L}_{cyc}(G, F) = \mathbb{E}_{x \in X}\left[\|\mathcal{F}(\mathcal{G}(x)) - x\|_1\right] + \mathbb{E}_{y \in Y}\left[\|\mathcal{G}(\mathcal{F}(y)) - y\|_1\right]. \quad (3)$$

Taking advantages of adversarial loss and cycle consistency loss, the model achieves a one-to-one correspondence mapping, and discovers the cross-domain relation [8]. The full objective is:

$$\begin{aligned}\mathcal{L}(\mathcal{G}, \mathcal{F}, D_X, D_Y) &= \mathcal{L}_{GAN}(\mathcal{G}, D_Y, X, Y) \\ &+ \mathcal{L}_{GAN}(\mathcal{F}, D_X, Y, X) + \lambda \mathcal{L}_{cyc}(\mathcal{G}, \mathcal{F}),\end{aligned} \quad (4)$$

where $\lambda$ controls the relative importance of the two objectives. However, the generative mapping functions $\mathcal{G}$ and $\mathcal{F}$ actually takes a dual responsibility for object transfiguration: to detect the objects of interest and to transfigure the object, which confuse the aims of the generative network.

On the other hand, we notice that the model can be viewed as two 'autoencoders': $\mathcal{F} \circ \mathcal{G} : X \to X$ and $\mathcal{G} \circ \mathcal{F} : Y \to Y$, where the translated image $\mathcal{G}(x)$ and $\mathcal{F}(y)$ can be viewed as intermediate representations trained by adversarial loss. In object transfiguration task, the generative mappings $\mathcal{G}$ and $\mathcal{F}$ are trained to generate objects to fool the discriminator. Therefore, the image background can be coded as any representation so long as it can be decoded back to the original, which does not guarantee background consistency before and after transformation. As a result, the proposed Attention-GAN that decomposes the generative network into two separate network: an attention network to predict the object of interests and a transformation network focuses on transforming object.

## 4 Model

The proposed model consists of three players: an attention network, a transformation network, and a discriminative network. The attention network predicts the region of interest from the original image $x$. The transformation network focuses on transforming the object from one domain to the other. The resulting image is therefore a combination of the transformed object and the background of original image with a layered operator. Finally, the discriminator aims to distinguish the real image $y \in Y$ and the generated image. The overview of the proposed model is illustrated in Figure 1(b). For notation simplicity, we only show the forward process that transforms images from domain $X$ to domain $Y$, and the backward process from domain $Y$ back to the domain $X$ can be easily obtained in the similar approach.

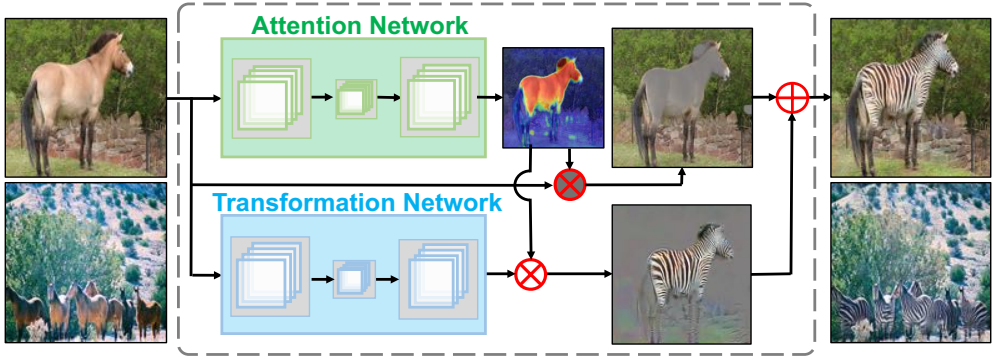

Figure 2: The proposed Attention-GAN for object transfiguration from one class to another. The attention network predicts the attention maps. The transformation network synthesizes the target object. A layered operation is applied on the background and transformed images to output the resulting image.

## 4.1 Formulations

The architecture of the proposed model is shown in Figure 2. Given an input image $x$ in domain $X$, the attention network $A_X$ outputs a spatial score map $A_X(x)$, whose size is the same as the original image $x$. The element value of score map is from 0 to 1. The attention network assigns higher scores of visual attention to the region of interest while suppressing background. In another branch, the transformation network $T$ outputs the transformed image $T(x)$ that looks similar to those in the target domain $Y$. Then we adopt a layered operation to construct the final image. Given transformed region $A_X(x)$, a transformed image $T_X(x)$ and image background from original image $x$ are combined as:

$$\mathcal{G}(x) \equiv A_X(x) \odot T_X(x) + (1 - A_X(x)) \odot x, \qquad (5)$$

where $\odot$ denotes the element-wise multiplication operator. Another mapping function $\mathcal{F}$ is introduced to bring transformed images $\mathcal{G}(x)$ back to the original space $\mathcal{F}(\mathcal{G}(x)) \approx x$. The mapping from an image $y$ in target domain $Y$ to the source domain follows:

$$\mathcal{F}(y) \equiv A_Y(y) \odot T_Y(y) + (1 - A_Y(y)) \odot y. \qquad (6)$$

Followed by Section 3, the *adversarial loss* (Equations (1) and (2)) and the *cycle consistency loss* (Equation (3)) are introduced to learn the overall mappings $\mathcal{G}$ and $\mathcal{F}$. In classical GANs [7, 8, 9], the generative mapping $\mathcal{G}$ transforms the whole image to target domain and then the generative mapping $\mathcal{F}$ is required to bring the transformed image back to original image $\mathcal{F}(\mathcal{G}(x)) \approx x$. However, in practice, the background of the generated image appears to be unreal and significantly different from the original image background, so that the cycle consistency loss can hardly reach 0. In our method, the attention network outputs a mask that separates the image into region of interest and background. The background part will not be transformed, so that the cycle consistency loss in the background reaches 0.

## 4.2 Attention Losses

Similar to cycle consistency, the attention map of object $x$ in domain $X$ predicted by attention network $A_X$ should be consistent with the attention map of the transformed object by attention network $A_Y$. For example, if a horse is transformed into a zebra, the region of the zebra should be brought back to the horse as a cycle. That is to say, the regions of interest in the original image and the transformed image should be the same: $A_X(x) \approx A_Y(\mathcal{G}(x))$. Similarly, for each image $y$ from domain $Y$, attention network $A_Y$ and $A_X$ should satisfy consistency: $A_Y(y) \approx A_X(\mathcal{F}(y))$. To that end, we propose an attention cycle-consistent loss:

$$\mathcal{L}_{A_{cyc}}(A_X, A_Y) = \mathbb{E}_{x \in X}\left[\|A_X(x) - A_Y(\mathcal{G}(x))\|_1\right] + \mathbb{E}_{y \in Y}\left[\|A_Y(y) - A_X(\mathcal{F}(y))\|_1\right] \tag{7}$$

In addition, we introduce a sparse loss to encourage the attention network to pay attention to a small region related to the object instead of the whole image:

$$\mathcal{L}_{A_{sparse}}(A_X, A_Y) = \mathbb{E}_{x \in X}\left[\|A_X(x)\|_1\right] + \mathbb{E}_{y \in Y}\left[\|A_Y(y)\|_1\right]. \tag{8}$$

Considering Equation (7), the attention maps of $A_X(\mathcal{F}(y))$ and $A_Y(\mathcal{G}(x))$ should be consistent to $A_Y(y)$ and $A_X(x)$, so they do not include additional sparse loss on $A_X(\mathcal{F}(y))$ and $A_Y(\mathcal{G}(x))$.

Hence, by combining Equations (1), (2), (3), (7) and (8), our full objective is:

$$\begin{aligned}\mathcal{L}(T_X, T_Y, D_X, D_Y, A_X, A_Y) &= \mathcal{L}_{GAN}(\mathcal{G}, D_Y, X, Y) + \mathcal{L}_{GAN}(\mathcal{F}, D_X, X, Y) \\ &+ \lambda_{cyc}\mathcal{L}_{cyc}(\mathcal{G}, \mathcal{F}) + \lambda_{A_{cyc}}\lambda L_{A_{cyc}}(A_X, A_Y) + \lambda_{A_{sparse}}\mathcal{L}_{A_{sparse}}(A_X, A_Y),\end{aligned} \tag{9}$$

where $\lambda_{attn}$ and $\lambda_{cyc}$ balance the relative importance of different terms. Attention network, transformation network and discriminative network in both $X$ domain and $Y$ domain can be solved in the following min-max game:

$$\arg \min_{T_X, T_Y, A_X, A_Y} \max_{D_X, D_Y} \mathcal{L}(T_X, T_Y, D_X, D_Y, A_X, A_Y), \tag{10}$$

The optimization algorithm is described in the Appendix.

## 4.3 Extra Supervision

In some cases, segmentation annotations can be collected and used as attention map. For example, our horse $\rightarrow$ zebra image segmentation of horse is exactly the region of interest. We therefore supervise the training of the attention network by segmentation label. Given a training set $\{(x_1, m_1), \cdots, (x_N, m_N)\}$ of $N$ examples, where $m_i$ indicates the binary labels of segmentation, we minimize the discrepancy between predicted attention maps $A(x_i)$ and segmentation label $m_i$. To learn the attention maps for both $X$ domain and $Y$ domain, the total attention loss can be written as:

$$\mathcal{L}_{A_{sup}}(A_X, A_Y) = \sum_{i=1}^{N_X} \|m_i - A_X(x_i)\|_1 + \sum_{j=1}^{N_Y} \|m_j - A_Y(y_j)\|_1. \quad (11)$$

The full objective thus becomes:

$$\begin{aligned}\mathcal{L}(T_X, T_Y, D_X, D_Y, A_X, A_Y) = & \mathcal{L}_{GAN}(\mathcal{G}, D_Y, X, Y) + \mathcal{L}_{GAN}(\mathcal{F}, D_X, X, Y) \\ & + \lambda_{cyc}\mathcal{L}_{cyc}(\mathcal{G}, \mathcal{F}) + \lambda_{A_{sup}}\mathcal{L}_{A_{sup}}(A_X, A_Y),\end{aligned} \quad (12)$$

where $\lambda_{cyc}$ and $\lambda_{A_{sup}}$ control the relative importance of the objectives. As the attention maps are supervised by semantic annotations, we do not incorporate the constraints of Equations (7) and (8).

## 5 Experiments

In this section, we first introduce two metrics to evaluate the quality of generated images. We then compare unsupervised Attention-GAN against state-of-the-art method. Next, we study the importance of the *attention sparse loss*, and compare our method against some variants. Lastly, we demonstrate empirical results of supervised Attention-GAN. Previously, the generative network for unpaired image-to-image translation [7, 8, 9, 10, 11] all undertake the dual responsibility: detecting and converting object. Considering CycleGAN achieves the state-of-the-art performance on unpaired object transfiguration and its implementation code is released by author[1], we choose CycleGAN as our compared method.

**Datasets.** We first evaluated the proposed Attention-GAN on three tasks: horse ↔ zebra, tiger ↔ leopard and apple ↔ orange. The images for horse, zebra, apple and orange were provided by CycleGAN [7]. The images for tiger and leopards are from ImageNet [12], which consists of 1,444 images for tiger, 1,396 for leopard. We randomly selected 60 images for test, and the rest for training. In supervised experiment, we performed horse ↔ zebra task where images and annotations can be obtained from MSCOCO dataset [32]. For each category, images in MSCOCO training set were used for training and those in MSCOCO val set were for testing. For all experiments, the training samples were first scaled as 286 × 286, and then randomly flipped and cropped as 256 × 256. In test phase, we scaled input images to the size of 256 × 256.

**Training Strategy.** For all experiments, the networks were trained with an initial learning rate of 0.0002 for the first 100 epoch and a linearly decaying rate that goes to zero over the next 100 epochs. We used the Adam solver [33] with batch size of 1. We updated the discriminative networks using a randomly selected sample from a buffer of previously generated images followed by [34]. The training process is shown in Appendix. The architectures of transformation networks and attention networks are based on Johnson *et al.* [35]. The discriminators are adapted from the Markovian Patch-GAN [36, 2, 7, 9]. Details are listed in the Appendix.

---

[1]CycleGAN [7] implementation code: https://junyanz.github.io/CycleGAN/

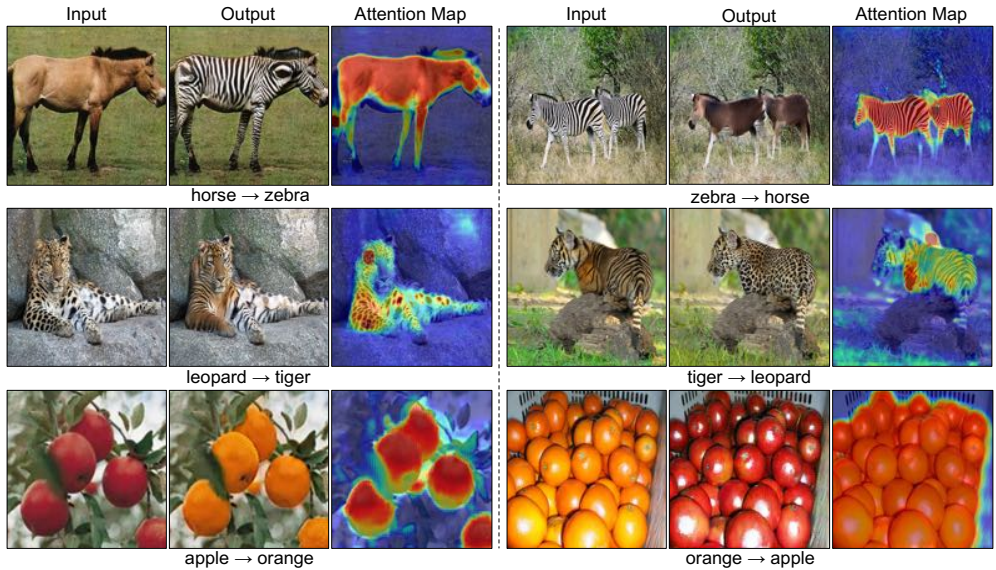

Figure 3: Results of object transfiguration on different tasks: horse ↔ zebra, leopard ↔ tiger and apple ↔ orange. In each case, the first image is the original images, the second image is the synthesized image, and the third image is the predicted attention map. Our proposed model only manipulates the attention parts of image and preserves the background consistency.

## 5.1 Assessment of Image Quality

Since object transfiguration is required to predict the region of interest and transform the object while preserve the background, we introduce metrics to estimate quality of transformed image.

To assess the background consistency of transformation, we compute PSNR and SSIM between generated image background and original image background. PSNR is an approximation to human perception of reconstruction quality, which is defined via mean squared error (MSE). Given a test sample $\{(x_1, m_1), \cdots, (x_N, m_N)\}$, we use pixel-wise multiplication $\odot$ by the segmentation mask to compute image background PSNR:

$$\frac{1}{N} \sum_{i=1}^{N} PSNR\left(x_i \odot (1 - m_i), \mathcal{G}(x_i) \odot (1 - m_i)\right), \qquad (13)$$

where $x_i$ is denoted as original image, $\mathcal{G}(x_i)$ is denoted as the resulting image, $(1-m_i)$ refers to the image background, so the pixel-wise multiplication $x_i \odot (1-m_i)$ indicates the background of original image, and $\mathcal{G}(x_i) \odot (1-m_i)$ indicates the background of generated image. Similarly, we use SSIM to assess the structural

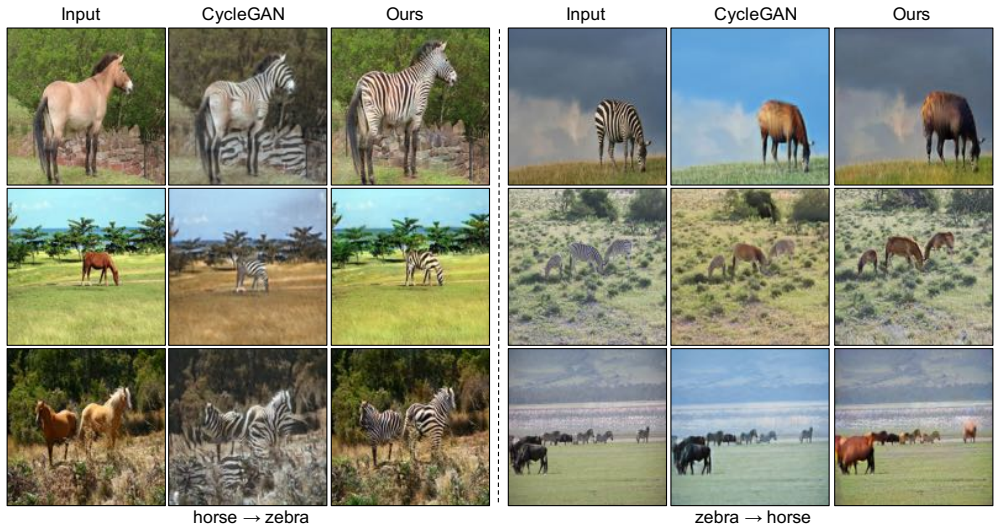

Figure 4: Comparison with CycleGAN on horse ↔ zebra. In each case, the first image is the input image, the second is the result of CycleGAN [7], and the third is the result of our Attention-GAN.

similarity between background of original image and composited output by using multiplication:

$$\frac{1}{N} \sum_{i=1}^{N} SSIM\left(x_i \odot (1-m_i), y_i \odot (1-m_i)\right). \qquad (14)$$

In experiment, we use test images and segmentation mask in the MSCOCO [32] dataset to evaluate background quality of generated image.

## 5.2 Unsupervised Results Comparisons to State-of-the-Art

### 5.2.1 Qualitative Comparison

Results of horse ↔ zebra are shown in Fig. 4. We observed that our method provides translation results of higher visual quality on test data than those of CycleGAN. For example, in the horse → zebra task, CycleGAN mistakes some parts of background as target and transforms them into black and white stripes. In the second column of Fig 4, CycleGAN translates the color of grass and trees from green into brown in the zebra → horse task. In contrast, our method generates zebra in the correct location and preserves background consistency. Comparison results on tiger ↔ leopard and apple ↔ orange are shown in Figure 5. The results of Attention-GAN are more visually pleasing than those of CycleGAN. In most cases, CycleGAN can not preserve background consistency, e.g., the color of jeans is transformed from blue to yellow in the first image, the color of water in third image is transformed from blue to yellow and the color of weeds in the last image is transformed from yellow to green. One possible reason is that our Attention-GAN disentangles the background and object of interests by the attention network and only transforms the object, while the compared method tends to use one generative network to manipulate the whole image.

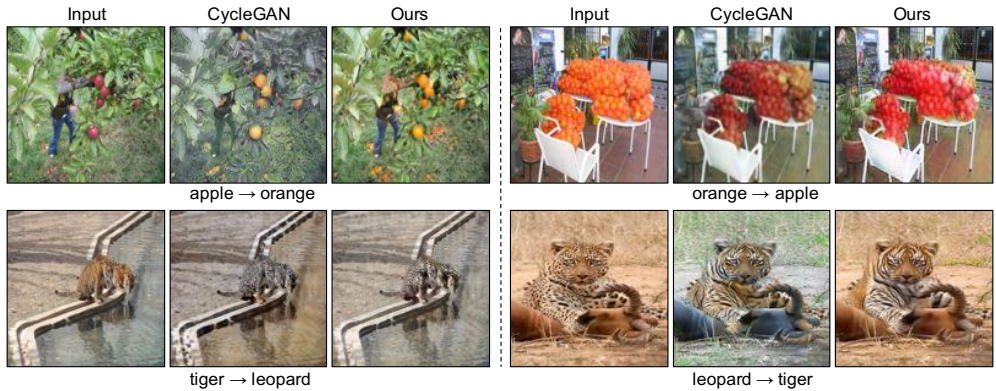

Figure 5: Comparison with CycleGAN on apple ↔ orange and tiger ↔ leopard. In each case: input image (left), result of CycleGAN [7] (middle), and result of our Attention-GAN (right).

### 5.2.2 Human Perceptual Study

We further evaluate our algorithm via a human subjective study. We perform pairwise A/B tests deployed on the Amazon Mechanical Turk (MTurk) platform. We follow the same experiment procedure as described in [37, 38]. The participants are asked to select the more realistic image from each pair. Each pair contains two images translated from the same source image by two different approaches. We test the tasks of horse ↔ zebra, tiger ↔ leopard and apple ↔ orange. In each task, we randomly select 100 images from test set. Each image are compared by 10 participants. Figure 6 shows the participants preference among 100 examples for each task. We observe that 92 results of our methods outperforms results of CycleGAN in horse ↔ zebra task while only one result of CycleGAN beats ours. In tiger ↔ leopard, still only 17% results of compared method beat ours, which indicates that qualitative assessments obtained by our proposed approaches are better than those obtained by existing methods. We also notice that in apple ↔ orange task, only 60 results of our methods outperform the compared method. We consider the reason is that a large portion of images in apple and orange dataset are close-up images whose background is simple so that CycleGAN could reach a competitive results with our method.

### 5.2.3 Quantitative Comparison

We evaluate the quantitative quality of object transfiguration by computing the PSNR and SSIM of image background (Equation (13) and Equation (14)). The test dataset is from MSCOCO dataset [32]. As MSCOCO dataset does not have the classes of tiger or leopard, and apples and oranges in images are too small, we only compare the results of horse $\leftrightarrow$ zebra. Results are shown in Table 1. As can be seen, for both PSNR and SSIM, our method in unsupervised fashion outperforms CycleGAN by a large margin, which indicates that the proposed model predicts accurate attention map and achieves a better performance of transformation quality.

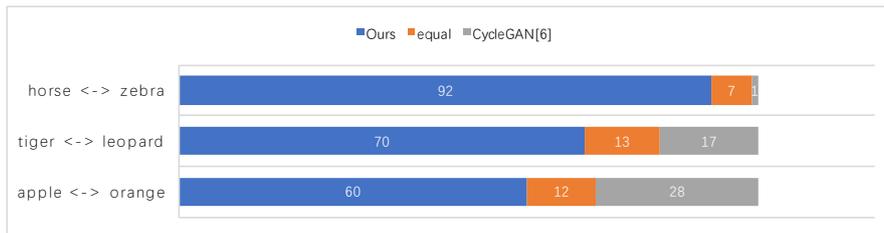

Figure 6: The stacked bar chart of participants preferences for our methods compared to CycleGAN [19]. The blue bar indicates the number of images that more participants prefer our results. The gray bar indicates the number of images that more participants prefer CycleGAN's results. The orange bar indicates the number of images where two methods get a equal number of votes from 10 participants.

Table 1: Background consistency performance of different object transfiguration tasks for background PSNR and SSIM.

|      | Task | CycleGAN | Ours (Unsupervised) | Ours (Supervised) |
|------|------|----------|---------------------|-------------------|
| PSNR | horse $\rightarrow$ zebra | 18.1875 | 22.2629 | 24.589 |
|      | zebra $\rightarrow$ horse | 18.1021 | 21.5360 | 23.9330 |
| SSIM | horse $\rightarrow$ zebra | 0.6725 | 0.9003 | 0.9482 |
|      | zebra $\rightarrow$ horse | 0.7155 | 0.8988 | 0.9534 |

### 5.3 Model Analysis

We perform model analysis on the horse $\rightarrow$ zebra task. Figure 7 shows the generated images, along with the intermediate generation results of model. In the second column, the attention maps show that the attention network of model is able to disentangle the objects of interests and the background from input image even in a completely unsupervised manner. The third column is the output of the transformation network, where the transformed zebra are visually pleasing while the background parts of images are meaningless. It demonstrates that the transformation network only focuses on transforming the object of interests. Moreover, Figure 7 shows that the final output images in the last column are combined by the background parts in the forth column and the objects of interests in the fifth column.

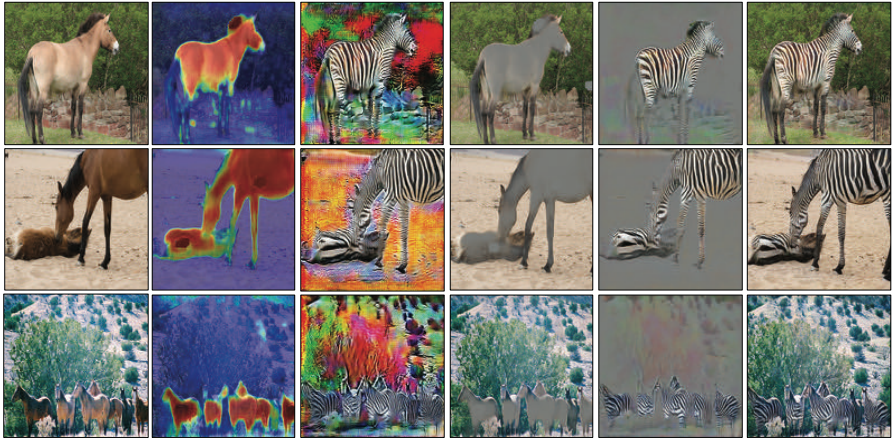

Figure 7: Generation results of our model on horse → zebra. From left to right: Inputs, attention maps, outputs of transformation network, background images factorized by attention maps, object of images factorized by attention maps, final composite images.

Table 2: Ablation study: performance of horse → zebra for different losses.

|  | $\lambda_{attn} = 0$ | $\lambda_{attn} = 1$ | $\lambda_{attn} = 5$ |
|---|---|---|---|
| PSNR | 19.8621 | 22.2629 | 24.2173 |
| SSIM | 0.8291 | 0.9003 | 0.9367 |

### 5.3.1 Ablation Analysis

In Figure 8, we show the qualitative results by different variants of our model on horse → zebra task. It can be seen that without the sparse loss ($\lambda_{attn} = 0$ in Equation (8)), the attention network would mistakenly predict some parts of image background as regions of interests. From PSNR and SSIM in Table 2, we observe that with the value of $\lambda_{attn}$ becoming larger, the performance of background consistency is better. However, when $\lambda_{attn}$ is set to be excess (e.g., $\lambda_{attn}$=5), we observe that in the last column of Figure 8 the attention mask shrank too much to cover the whole object of interests, which decreases the qualities of transformed object as shown in the penultimate column of Figure 8). It is because if we emphasize too much on the relative importance of sparse loss in the full objective (Equation (4)), the attention network can not comprehensively predict the object location. We find $\lambda_{attn} = 1$ is an appropriate choice, which makes a good balance to pay enough attention to the objects of interests.

## 5.4 Comparisons of Supervised Results

We compute PSRN, SSIM of background region between generated and original images in horse ↔ zebra task. As shown in Table 1, the Attention-GAN with supervision outperforms unsupervised Attention-GAN and CycleGAN from the perspective of background consistency over the whole test dataset. This demonstrates that with the segmentation mask, the attention network predicts the object of interests more accurately. Figure 9 shows qualitative results of the proposed model with supervised attention information. In contrast, the compared methods predict some parts of the person as target object and transform them into black and white stripes. For instance, in the first row of Figure 9, the faces of man are transformed into black and white stripes by CycleGAN and unsupervised Attention-GAN. We also notice that the attention maps with supervision tend to be dark red or dark blue, which indicates the supervised at-tention network detects the object with high confidence, and the attention mask disentangles the background and object of interests more clearly.

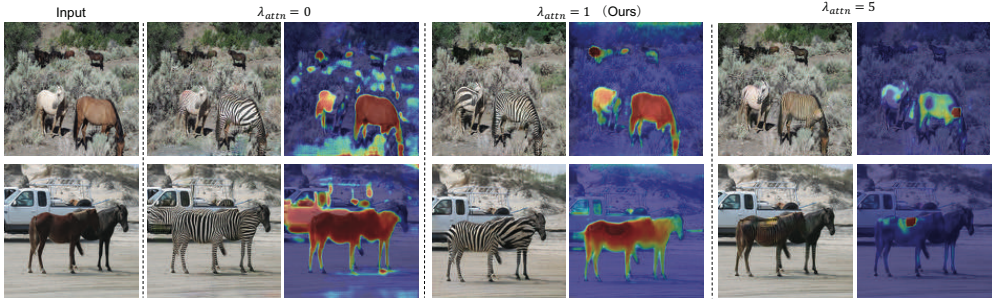

Figure 8: The effect of sparse loss with different parameters $\lambda_{attn}$ for mapping horse $\rightarrow$ zebra. From left to right: input, output and attention map without sparse loss, input and attention map when $\lambda_{attn} = 1$, input and attention map when $\lambda_{attn}=5$.

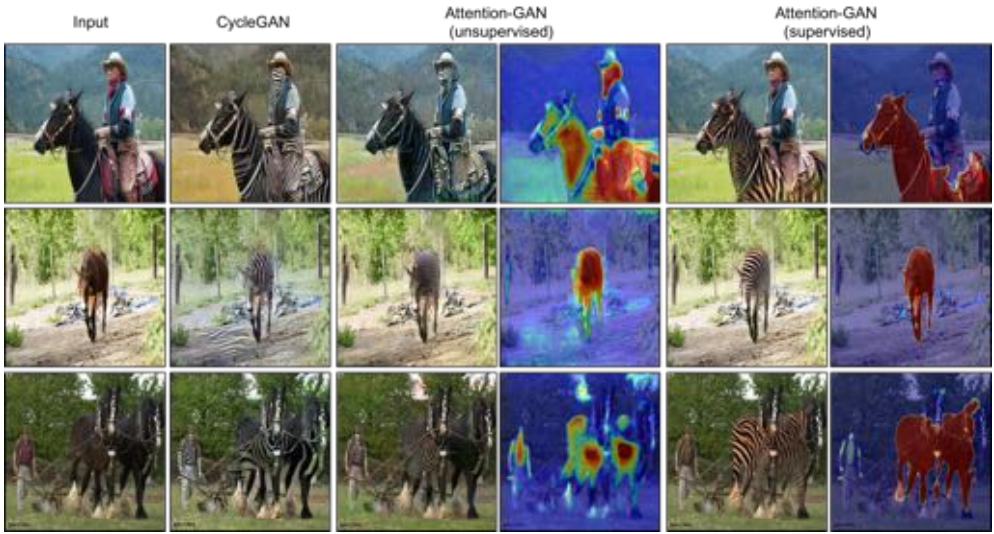

Figure 9: Comparison of horse $\leftrightarrow$ zebra between supervised Attention-GAN (right), CycleGAN [7] (left) and unsupervised Attention-GAN (middle).

# 6 Conclusion

This paper introduces attention mechanism into the generative adversarial networks on object transfiguration. Different from classical GANs whose generative network undertakes two responsibilities, we develop a three-player model that consists of an attention network, a transformation network and a discriminative network. The attention network predicts the regions of interest whilst the transformation network transforms the object from one domain to another. To guarantee the attention can be paid to object, attentional cycle consistency loss and sparse loss are introduced. The results demonstrate that the necessity of investigating attention and that the proposed algorithm produces high quality and practically preferred results for object transfiguration.